# Existence of Hierarchies and Human's Pursuit of Top Hierarchy Lead to Power Law

Shuiyuan Yu[1], Junying Liang[2], Haitao Liu[2*]

**Abstract**

The power law is ubiquitous in natural and social phenomena, and is considered as a universal relationship between the frequency and its rank for diverse social systems. However, a general model is still lacking to interpret why these seemingly unrelated systems share great similarity. Through a detailed analysis of natural language texts and simulation experiments based on the proposed 'Hierarchical Selection Model', we found that the existence of hierarchies and human's pursuit of top hierarchy lead to the power law. Further, the power law is a statistical and emergent performance of hierarchies, and it is the universality of hierarchies that contributes to the ubiquity of the power law.

**Keywords:** power law, hierarchy, exponent of power law, frequency rank, Zipf's law

## 1   Introduction

The frequency and its rank of numerous natural and man-made phenomena can be described as the power-law relationship $p(r) \sim r^{-\alpha}$, where p is the frequency of the occurrence, r indicates the frequency rank, and the constant α is referred to as the power-law exponent. Research has documented a large number of phenomena that follow the power law, such as the city populations [1,2], individual wealth [3,4], firm sizes [5], the frequency of occurrence of personal names [6], the number of citations received by papers [7], the scale of war [8], the sales volume of books, music disks [9], market fluctuations[10], moon craters [11], the numbers of species in biological taxa [12], energy released in earthquakes [13] and so on [a review see 14]. According to incomplete statistics, about 800 articles and monographs have been published [15] by June 2011. Among all different interpretations of the relationship, of great fame is Zipf's law [1], theorizing that in texts of human language, the frequency of a word has an inverse proportion to its rank.

Approaches differ in this line of research. Some studies have investigated particular phenomena associated with specific areas [16-21], whereas others propose several mechanisms to account for these phenomena [22-25]. However, most of these mechanisms have only introduced an abstract model, without providing solid explanations related with general phenomena. Moreover, some fundamental questions still remain open for discussion. One key question is what kind of universality is reflected in the power law. Specifically, does the frequency rank in the power law only represent the rank of the frequency, or is it related with the meaning of the power law? What is the significance of the power-law exponent α? What factors contribute to the change of the power-law exponent?

[1]  School of Computer, Communication University of China, Beijing, 100024, P.R. China. [2] Department of Linguistics, Zhejiang University, Hangzhou, 310058, P.R. China.

* To whom correspondence should be addressed. E-mail: htliu@163.com

To explore the above questions, this study proposed the Hierarchical Selection Model by examining a large pool of natural language texts. To anticipate, this model can explain the meaning of the power law exponent and almost all of the frequency-rank relationships, and provide a controlling method for the power law system. These explanations have gained support from previous research, and the predications of this model are broadly in line with the natural and social phenomena. We found that for diverse phenomena in human society, the existence of hierarchies and human's preference for top hierarchy lead to the power law. In other words, the power law is a statistical and emergent performance of hierarchies. Given that "It is a commonplace observation that nature loves hierarchies" [26], the power law also possesses the characteristics of popularity and universality.

## 2  Text analysis and the linguistic meaning of power law

Words in text are sequenced according to rules and purposes. Some words, such as function words, can appear in texts of any topic, whereas many content words only appear in texts of associated topics, suggesting that the distribution frequency of a word is associated with topics it can appear in. It is likely that the frequency rank is also correlated with topics. In the present study, we chose short news reports with different topics as the pool of texts data in order to explore the influence of topic number that a word can appear in upon the word frequency distribution, and to further investigate the significance of word frequency rank.

The pool of texts data is composed of 10,287,433 word tokens, with 138,243 word types. The texts include 8 topic subsets, namely, IT, health, sports, tourism, education, employment, history and military. Each subset contains about 1990 short Chinese news reports. Table 1 is an overview of the texts and the eight topics, specifically, the word number, cumulative word frequency in topics, and their respective power-law fitness results between the word frequency and its rank of each topic and the collection of all texts. Here, the curve fitting was obtained by using the method of least squares. The results include the goodness of fit (Adjusted $R^2$) and the power law coefficients. This method was adopted in the following power-law fit. Figure 1 depicts the curves of log-log coordinate of the word frequency of total texts and texts of each topic with their corresponding frequency rank.

**Table 1** A Summary of Statistics of Eight topics and All Texts

| Topics | Word | Word Frequency | Power-law Fitness | |
|---|---|---|---|---|
| | | | Exponent | Goodness (Adjusted $R^2$) |
| IT | 30, 060 | 860, 344 | -1.491 | 0.9732 |
| Health | 38, 854 | 1, 300, 509 | -1.532 | 0.9687 |
| Sports | 24, 968 | 717, 225 | -1.485 | 0.9713 |
| Tourism | 46, 513 | 1, 009, 260 | -1.4 | 0.9722 |
| Education | 44, 667 | 1, 586, 666 | -1.521 | 0.9769 |
| Employment | 37, 077 | 1, 422, 806 | -1.519 | 0.9752 |
| History | 70, 968 | 2, 170, 285 | -1.464 | 0.9704 |
| Military | 37, 132 | 1, 220, 338 | -1.532 | 0.9716 |
| Collection | 138, 243 | 10, 287, 433 | -1.705 | 0.9716 |

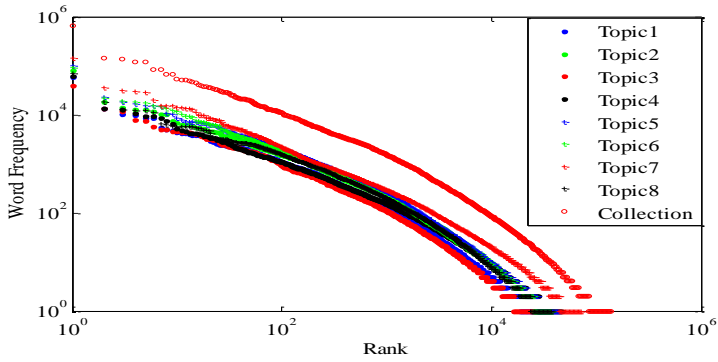

**Fig. 1**. The log-log coordination curves of the relationship between the word frequency and its rank in texts of 8 topics and texts collection.

The relationships between the word frequency and its rank in topics texts and texts collection both follow the power law, with the exponent α close to 1.5. However, the exponent of the collection is around 1.7, evidently higher than all the subsets.

Words have their own distinctive contexts of occurrence in that some words appear in contexts of any topic, and are generally high in word frequency but low in word frequency rank; whereas some other words appear only in a few topics, and in general have a low word frequency but a high frequency rank. Apparently, the frequency rank of a word is associated with the number of topics this word appears in. As shown in Figure 2, the horizontal axis is the word frequency rank, and the vertical axis is the probability density. Eight curves represent the number distribution of topics a word belongs to (hereinafter referred to as 'the number of topics', abbreviated as 'NT'). Curve 8 indicates that this word appears in all topics, regardless of any specific topic. Curve 1 indicates that this word is only used in one topic, a highly topic-specific word. As plotted is the estimated density of the number of topics for a word using a kernel smoothing method [27]. Clearly, the number of words in each field concentrated on the locations reversely ordered in the rank of word frequency, implicating that the rank of word frequency is correlated with the NT.

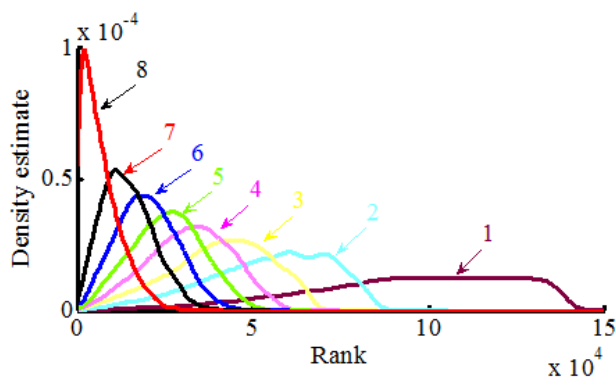

**Fig. 2.** The estimated density function of the number of topics distributing in word frequency rank. Eight distributions of NT are in sequence on the axis of Rank, indicating that the frequency rank also has the meaning of NT.

To examine the nature of the NT, we calculated the number of words in each field, the average word frequency rank, and the power-law fitting results of word frequency distribution in each field.

**Table 2** The distribution of various types of word. The meaning of the second column (Average word frequency rank), as previously mentioned in Fig.2, is that the average word frequency rank of each NT has a monotonic sequence, suggesting that there exists a correlation between NT and the rank; The third column (Number of words) indicates that NT has an inversed relationship with the number of words this NT has. The fifth column (Word frequency percentage) reveals that most words constituting a text are chosen from the NT=8 subset. The last two columns manifest that the word frequency in each NT follows the power law quite well.

| The NT | Average word frequency rank | Number of word | Word number percentage | Word frequency percentage | Average word frequency | Power-law fitting results of word frequency distribution | |
|---|---|---|---|---|---|---|---|
| | | | | | | Exponent | Goodness (Adjusted $R^2$) |
| 1 | 94228.12 | 81864 | 59.22 | 2.22 | 2.79 | -0.735 | 0.9327 |
| 2 | 54303.72 | 16156 | 11.69 | 1.27 | 8.06 | -0.8456 | 0.9798 |
| 3 | 40522.58 | 9603 | 6.95 | 1.44 | 15.41 | -0.86 | 0.9768 |
| 4 | 31883.32 | 6687 | 4.84 | 1.50 | 23.10 | -0.8116 | 0.9656 |
| 5 | 25438.21 | 5546 | 4.01 | 1.89 | 35.05 | -0.8088 | 0.9568 |
| 6 | 19734.64 | 4672 | 3.38 | 2.63 | 57.9 | -0.8195 | 0.9489 |
| 7 | 14205.00 | 4984 | 3.61 | 5.18 | 106.91 | -0.8941 | 0.9353 |
| 8 | 6641.88 | 8731 | 6.32 | 83.87 | 988.25 | -1.337 | 0.9258 |

With the increase of the NT, the number of words is monotonically decreasing, but begins to rise when the NT reaches 8 (Table 2), thus demonstrating a U-shaped distribution. Further, we used two power function fits for the relationship between the number of words and the NT. The fitting function is $f(x) = 81530x^{-2.094} + 69.9 x^{2.26}$, where x represents the NT, f(x) is the number of words with NT as x. As seen from Table 2 and the fitting function, two rules affect the number of words in a field: The first is a decreasing rule, mainly relevant to the specific topic-related words occurring in a small NT, that with the increase of the topic availability, or the applicability of the word, the number of words significantly decreases in a manner of power-law. The second is an increasing rule, mainly relevant to the topic-free words, that when the use of word is extending to any field, the number of words once again increases in the manner of power-law.

To further analyze the relation between the word frequency rank and the NT, we computed their correlation coefficient. The number of words in each field obeys a power-law, so computing is on the same proportional number rather than the same number of words. The resulting correlational coefficient is -0.9811, suggesting that the frequency rank of a word is correlated with its NT.

Next, we analyzed the distribution of cumulative word frequency in NT in Table 2. While composing a text, nearly 84% words have the NT as 8, and put differently, are common words that can be used in any field. Instead, words with the NT 1~6, the frequency ratios are about 1-2 percent. For the NT as 7, the word frequency is slightly higher than that of the NT 1~6. The power-law fitting results for the word frequency percentage is $f(x) = 83.84(9-x)^{-3.791}$, with goodness adjusted $R^2 = 0.9971$, where x represents the NT, and f is the word frequency percentage. It is worth noting that the group of words with the smallest number (namely, with the NT as 8) gets the highest word frequency percentage.

In addition, among words with a same NT, only a small number of words have an absolute high word frequency, whereas the majority of words have a very low frequency, the fitting results of which still follow the power law.

The distribution of word frequency in each field (As shown in Table 2 Column 5) indicates when we conceptualize what we intend to say, we are most likely to choose common words that can be used in all topics, and then a few content words help flesh out, making the expression specific and significant. Grammatically presented, all the common words can be taken as formulating a construction, and then the slots in this construction will be filled up with content words, thus forming a complete exposition, a notion that is consistent with the major tenets of Construction Grammar[28-30]. Given that the NT of a word represents how many topics this word can appear in, it per se means the word's 'expressive ability'. As shown in Table 2, the higher the words' 'ability', the fewer the words. That is, words are divided into different hierarchies. Albeit that the number of words at the highest hierarchy is the smallest, they have the highest probability of being selected, and vice versa.

The data pattern here suggests that in human language, each word has its own NT, and the relationship between the number of words and the NT that words can appear in follows the power law. When people choose words to express ideas, most likely they will use words that are not limited by topics, indicating that word frequency and the degree of limit have a power-law relationship. That is to say, the word frequency-rank is an inherent attribute of human languages, instead of being determined by any text. Taken together, it is the inherent word frequency rank of a language and these several power laws that generate the power law between word frequency and its rank.

## 3   Hierarchical Selection Model and its simulation results

As displayed in Fig. 2 and Table 2, the frequency of a word is determined by the following two parameters: One is the NT this word can appear in, and the other is its probability of being selected. According to the computation of conditional probability, while choosing words to compose texts, the formation of word frequency can be divided into two steps: choose the NT, and then choose words among those with the same NT. In this vein, there are three factors affecting the relationship between the word frequency and its rank: First, the distribution of word numbers in NT $f_m(x)$, where x refers to the NT; Second, the distribution of word frequency in each NT $f_w(x)$, where x is the frequency rank of words with the same NT; Third, each field's probability of being selected $f_c(x)$, where x represents the NT.

To explore the generalizability of the power law, we propose a model from the above findings, named as "Hierarchical Selection Model":

> Given there are N 'objects', (the equivalent of words), belonging to M hierarchies (The hierarchy number is equivalent to the NT of words), $M \ll N$;
> 
> The numbers of 'objects' at different hierarchies have the distribution $f_m(x)$, where x is the hierarchy number;
> 
> When a hierarchy is selected, for each 'object' in it, the probability of being selected has distribution $f_w(x)$, where x refers to the frequency rank of words at that hierarchy. During every selection, the frequency rank of each word remains unchanged.
> 
> For each hierarchy, the probability of being selected has distribution $f_c(x)$, where x is the hierarchy.

The process of selecting objects follows two steps: first selecting the hierarchy, and then selecting the objects within this hierarchy.

For this Hierarchical Selection Model, of our interest is the limit distribution of each object's frequency after repeated selections. We assume that $f_w$ at each hierarchy is fixed, since $f_m$, $f_w$ and $f_c$ all show a power-law distribution, the resulting power-law is hence the superposition of these power-laws. Does the relationship between word frequency and its rank continue to follow the power law if the three distribution functions have much slower changing forms than the power law? To this end, we used the triangular distribution on the $f_m$, $f_w$ and $f_c$ for simulation, wherein the total number of objects was kept constant, and the number of hierarchies and the ratio of maximum and minimum values were changed systematically in these triangular distributions. The goodness of fitting the frequency- rank relationship of objects to power law and the power-law exponent were computed.

The analysis of variance (ANOVA) of the fitting goodness with respect to the number of hierarchy, $f_m$, $f_w$, and $f_c$ show that all of their p-values are 0.000, indicating that the four factors all have a significant effect on the fitting goodness of the power law. When the ratio between the maximum and minimum number of objects at the hierarchies is greater than 3, as long as the $f_c$ is not a uniform distribution, the resulting relationship between the frequency and its rank will follow a power-law distribution.

Of particular note is that when $f_w$ has a uniform distribution, the result can still follow the power law. Figure 3 shows the contour of $f_m$ and $f_c$ on the fitting goodness when $f_w$ has the uniform distribution and the number of hierarchy is 5. In many conditions, the result can be a power law. Figure 4 is the curves of three simulation results. It manifests the relation of the object's frequency of being chosen and its rank. The linear fitness is all good after logarithm (the fitting goodness are 0.8926, 0.9146 and 0.9275 respectively).

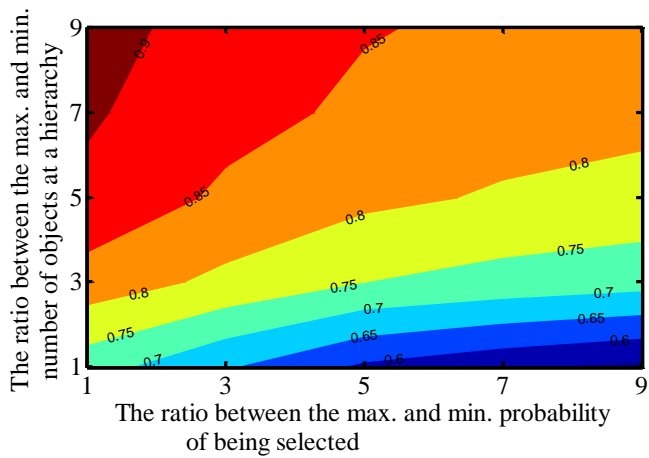

**Fig. 3.** The simulation results of triangular distribution. The contour of fitting goodness on $f_m$ and $f_c$ when the number $f_w$ has the uniform distribution and the number of hierarchy is 5. Values in this figure are the fitting goodness of power law. Data at the upper left part show that in this condition, the results still show a power-law relationship. This indicates that, even if objects at each hierarchy are evenly selected, when the number of objects differs across hierarchies, and the hierarchy with a smaller number of objects has a bigger probability of being selected, the frequency-rank result is still in line with the power-law relationship.

In order to examine the impact of the number of hierarchies, $f_m$, $f_w$, and $f_c$ on the power-law exponent $α$, the ANOVA of the power law exponent with respect to the four factors is computed. All p-values are 0.000, indicating that each of them has a significant effect on the power law exponent. The regression analysis of power law exponent exhibits that $f_m$, $f_w$ and $f_c$ all have a positive correlation with $α$. That is, the greater the degree of non-uniform distribution, the greater its power-law exponent α.

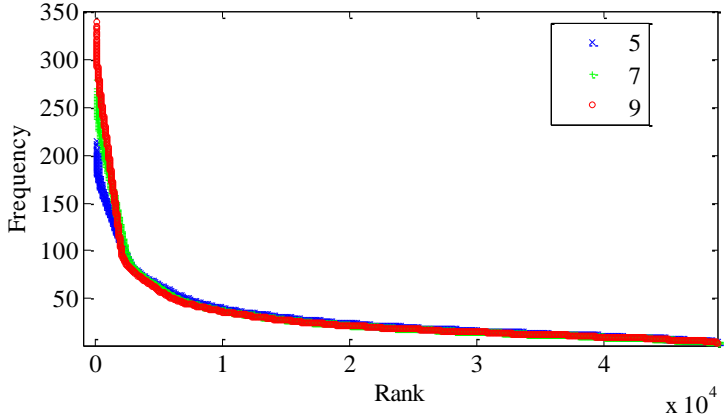

**Fig. 4.** Simulation results of the triangular distribution. When the number of hierarchy is 5, the number of objects is 50000, and the number of repeated selection reaches 2,000,000, the object's probability of being selected within each hierarchy will be uniformly distributed, and the number of objects within each hierarchy and the hierarchy's probability of being selected are both subject to the simulation of triangular distributions. The numbers 5, 7, 9 in this figure are the ratios between the maximum and the minimum number of objects within each hierarchy.

The experimental results suggest that when the objects are divided into hierarchies, the number of objects in each hierarchy varies widely, and the hierarchy with a smaller number of objects have a larger probability of being selected, even if the objects within a same hierarchy has an equal probability of being selected, the resulting frequency-rank relationship can still follow the power law.  If the hierarchy with the smallest number of objects and selection priority is taken as the highest one (which is same as many social and natural phenomenon), the mechanism for the generating of power law can be attributed to the existence of hierarchies. We thus suggest that the existence of hierarchies and the preference for top hierarchy lead to the power law. Previous research on the non-linear growth of complex network [31] suggests that, when the number of network nodes increases, if the correspondingly increasing edges have the priority linear growth (the same as the triangular distribution in this article), then the degree of nodes follows the power-law distribution, which is consistent with the conclusion in the current research. Importantly, our results not only accounts for the reasons why the degree of nodes in the complex network follows the power-law, but also may provide a solid explanation for the causes of power laws in the human society.

Further, we explored the relationship between the exponent $\alpha$ and the number of hierarchies, the number of objects based on the Hierarchy Selection Model. In the simulation experiment, $f_m$, $f_w$ and $f_c$ were fixed as the exponential distributions, the exponential values of their distributions were taken from the corresponding values in Table 2. The number of objects and the number of hierarchies were changed systematically, and then we fitted the relationship between the object's frequency and its rank to power law and computed the exponent. The mean of fit goodness is 0.9363 and the standard deviation is 0.0204, indicating that the frequency-rank relationship of objects obeys the power law. The relationship between the power law exponent and the number of objects (Fig. 5A) suggests that when the number of hierarchies remain unchanged, the more the number of objects, the smaller the exponent α; Moreover, the relationship between the power law exponent and the number of hierarchies (Fig. 5B) suggests that when the number of objects remains unchanged, the more the hierarchies, the bigger the exponent α will be. In short, the power law exponent manifests the bias degree of "abilities" that all the objects in the system have.

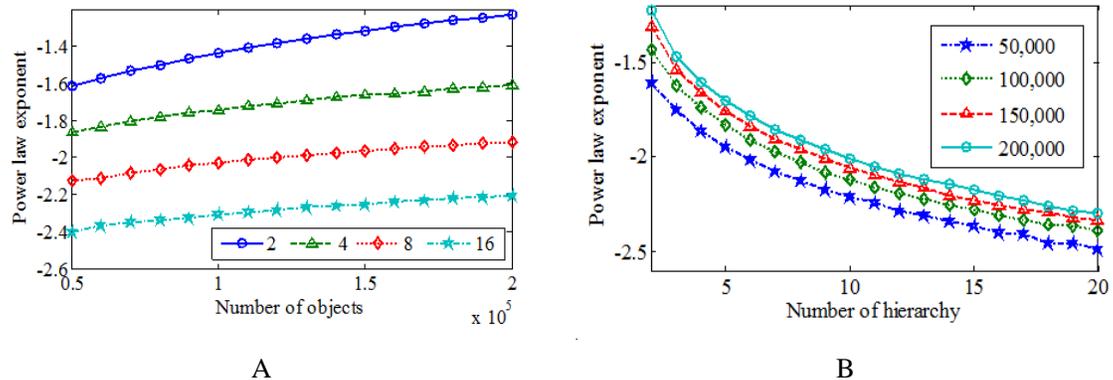

A                                                                 B

**Fig. 5. A.** The power-law exponent as the monotonic function of the number of objects. Each curve represents the *α* of various numbers of hierarchies. That is, when the number of hierarchies keeps unchanged, the bigger the number of objects, the smaller the power-law index α, and vice versa. **B.** The power-law exponent as the monotonic function the number of hierarchies, each curve represents the *α* of various numbers of objects. That is, when the number of objects keeps unchanged, the more the hierarchies, the bigger the power-law exponent *α*, and vice versa.

If the topic of a discourse has changed during the information flow, the simulation results are effective in predicting the changes in the exponents: the beginning of a new topic does not significantly increase the number of words (Table 2 shows the majority of words in the new topic are topic-free words, that is, the same words with the original topic), but heightens the number of hierarchies, which increases the power law exponent *α* (Fig. 5B). If a speech contains many different topics, then its power law exponent will rise significantly. For example, the language of the fragmented discourse schizophrenia has such features, thus their power law exponent *α* is relatively higher [32]. A large text generally involves a diversity of topics, such that it has a larger power-law exponent. This is congruent with the predictions of the "Hierarchical Selection Model", and also broadly in line with the data in Table 1.

## 4　The applications of Hierarchical Selection Model

Among most of the already found power-law phenomena in nature or society, one property is shared: there exist hierarchies, and the top hierarchy is always preferentially selected. For instance, a couple of cities in one country, with better conditions and resources, are widely preferred as ideal places for career and inhabitation; Likewise, a small number of journal articles, with their highest academic values, are frequently quoted by scientists in various fields of research. The simulation results based on our model suggest that the existence of hierarchical structures and human's preference for top hierarchy lead to the power law. Table 3 lists the power-law phenomena with hierarchical structures, and their explanations on the basis of the Hierarchical Selection Model. The model elements and the corresponding phenomena are also included.

**Table 3** Explanations for the power-law phenomena based on 'Hierarchical Selection Model'. In some phenomena, neither of the existence of 'hierarchies' or human's 'selection' is explicit. Listed in this table are the relationships between the elements of our model and their counterparts in the power-law phenomena, which can explain the causes of the power law.

| Phenomenon | 'Object' | Meaning of 'Hierarchy' | Implication of 'Selection' |
|---|---|---|---|
| City population [1, 2] | City | city's international reputation, work and promotion opportunities, wage levels, urban construction and other personal choices | Personal choice of city life |
| War [8] | War | the number of countries involved in wars, war interests | The number of wars per capita |
| Name [6] | Name | The meaning of a name, a former celebrity bearing the same name, culture-orientation | Naming |
| The number of scientists' papers[7] | scientists | Research ability, research fields, popularity of research questions | Writing every single paper |
| Web-page click-throughs[33] | Web-page | Fields of content，information of each field | Click-throughs |
| Sales of books and music records[9] | Books and music | Popularity of content | Every purchase |
| Firm size[5] | capital | Funds, personnel, patents, brands and etc. | Flow of each fund |
| Annual personal income[3, 4] | persons | Value of labor and social resources | allocation of resources every time |

As it can be seen, the number of objects at each hierarchy is unequal, the higher the hierarchy, the smaller the number of objects. Further, since man's preference for excellence is in accord with the ordinal level of hierarchies, and thus the higher the hierarchy, the higher the probability of being selected for the objects at that hierarchy. This account satisfies the conditions for the generating of power law suggested by the 'Hierarchical Selection Model'.

The existence of hierarchies and the pursuit of top hierarchy lead to the power law in various natural and social phenomena. This suggests, albeit it might not be the fundamental mechanism for the generating of power law, compared with preexisting models, the Hierarchical Selection Model enjoys more universal and explanatory power.

## 5  Discussion and conclusion

The word frequency and its rank in natural texts follow the power law (in special cases referred as Zipf's Law), and its underlying mechanism, as well as that of some social phenomena, is suggested as follows: 'objects' are hierarchical, such that each hierarchy represents the object's 'ability' or comprehensive power of 'abilities', and different hierarchy contains a different number of 'objects', the hierarchy with a smaller number of objects has a bigger probability of being selected. Essentially, the power law exponent represents the bias level of object's abilities, and a statistical performance of hierarchies.

As the power law means a non-uniform distribution, an excessive degree of non-uniformity may be largely undesirable in some circumstances, such as the over-concentration of urban population, monopolization from several large companies, discrepancy of personal wealth. Then how to manipulate the degree of non-uniformity, that is, the power law exponent $\alpha$? Simulation results in the present study point out that some fundamental measures can be taken to eliminate the power laws or decrease the power law exponent $\alpha$, for instance, reducing the gap between objects' probability of being selected at a same hierarchy, or decreasing the gap between hierarchies' probability of being selected so as to undermine man's irrational pursuit of high hierarchy by diversifying the value orientations. Take the over concentration of urban population for an example, the "power" of cosmopolitans at the highest hierarchy can be divided, such that the economic, political, cultural, and transportation centers will be scattered to different cities, whereas the cities at a lower hierarchy can be diversified by adding natural and cultural charms, decreasing the living cost and some others, to weaken man's single desire for the highest hierarchy.

We caution that for some other scale-rank relationships, such as the earthquake size, the moon craters, solar flares, the number of species and etc, the 'Hierarchical Selection Model' is still rather under-qualified to provide explanations, and needs to be further developed.

To conclude, the power law is generated due to the ability-based hierarchies of objects, and human's single pursuit of highest hierarchy. In human society, the social order and value system give birth to the general hierarchical structures, and facilitate man's constant and consistent pursuit of highest hierarchy, such that the power laws are ubiquitous in the world, particularly in human society.

**References**

[1]   Zipf G K. Human Behaviour and the Principle of Least Effort. Addison-Wesley, Reading, MA, 1949

[2]   Hernan A Makse, Shlomo Havlin, H Eugene Stanley. Modelling urban growth patterns. Nature, 1995, 377: 608-612

[3]   Simon H A, Bonini, C P. The size distribution of business firms. American Economic Review, 1958, 48: 607-617

[4]     Chatterjee A, Chakrabarti B K. Kinetic exchange models for income and wealth distributions. The European Physical Journal B, 2007, 60: 135–149

[5]     Robert L Axtell. Zipf distribution of US firm sizes. Science, 2001, 293(5536): 1818-1820

[6]     Zanette D H, Manrubia S C. Vertical transmission of culture and the distribution of family names. Physica A, 2001, 295: 1-8

[7]     Price D J de S. Networks of scientific papers. Science, 1965, 149: 510-515

[8]     Roberts D C, Turcotte D L. Fractality and Self-Organized Criticality in Wars. Fractals, 1998, 4: 351-357

[9]     Redner S. How popular is your paper? An empirical study of the citation distribution. Eur. Phys. J. B, 1998, 4: 131–134

[10]    Gabaix X, Gopikrishnan P, Plerou V, Stanley H E. A theory of power-law distributions in financial market fluctuations. Nature, 2003, 423: 267-270

[11]    Neukum G, Ivanov B A. Crater size distributions and impact probabilities on Earth from lunar, terrestialplanet, and asteroid cratering data. In Gehrels T(ed.), Hazards Due to Comets and Asteroids, pp. 359–416, University of Arizona Press, Tucson, AZ, 1994

[12]    Willis J C, Yule G U. Some statistics of evolution and geographical distribution in plants and animals, and their significance. Nature, 1922, 109: 177–179

[13]    Sornette D, Knopoff L, Kagan Y Y, Vanneste C. Rank-ordering statistics of extreme events: Application to distribution of large earthquakes. J Geophys Res, 1996, 101: 13883–13893

[14]    Mitzenmacher M. A brief history of generative models for power law and lognormal distributions. Internet Mathematics, 2004, 1: 226–251

[15]    http://www.nslij-genetics.org/wli/zipf/

[16]    Kello CT, Brown GDA, Ferrer-i-Cancho R, Holden G, Linkenkaer-Hansen K, Rhodes T, Orden GC Van. Scaling laws in cognitive sciences. Trends in Cognitive Sciences, 2010, 14(5): 223-232

[17]    Popescu I-I, Altmann G, Köhler R. Zipf's law - another view. Quality and Quantity, 2010, 44: 713-731

[18]    Gustavo Martínez-Mekler, Roberto Alvarez Martínez, Manuel Beltrán del Río, Ricardo Mansilla, Pedro Miramontes, and Germinal Cocho. Universality of rank-ordering distributions in the arts and sciences. PLoS ONE, 2009, 4(3):e4791

[19]    Xavier Gabaix, Parameswaran Gopikrishnan, Vasiliki Plerou, Eugene Stanley H. A theory of power-law distributions in financial market fluctuations. Nature, 2003, 423: 267-270

[20]    West G B, Brown J H, Enquist B J. A general model for the origin of allometric scaling laws in biology. Science, 1997, 276: 122-126

[21]    Tuzzi A, Popescu I.-I., Altmann G. Zipf´s law in Italian texts. Journal of Quantitative Linguistics, 2009, 16(4): 354-367

[22]    Seung Ki Baek, Sebastian Bernhardsson, Petter Minnhagen. Zipf's law unzipped. New Journal of Physics, 2011, 13: 043004

[23]    Corominas-Murtra B, Sole R. Universality of Zipf's law. Physical Review E, 2010, 82: 011102

[24]    Newman N. The power of design. Nature, 2000, 405: 412-413

[25]    Popescu I-I, Altmann G. Zipf´s mean and language typology. Glottometrics, 2008, 16: 31-37

[26]    Simon H. The organization of complex systems, in Pattee H(ed.). Hierarchy Theory: The Challenge of Complex Systems (NY: George Braziller). pp. 1-27, 1973

[27]    Bowman A W, Azzalini A. Applied Smoothing Techniques for Data Analysis. Oxford University Press, 1997

[28]　Fillmore C, Kay P, O'Connor M C. Regularity and idiomaticity in grammatical constructions: The case of let alone. Language, 1988, 64: 501-538

[29]　Kay P, Fillmore J. Grammatical constructions and linguistic generalizations: The what's X doing Y constructions. Language, 1999, 75: 1-33

[30]　Goldberg A. Constructions: A Constructional Approach to Argument Structure. Chicago: Chicago University Press, 1995

[31]　Albert R, Barabási A.-L. Statistical mechanics of complex networks. Reviews of Modern Physics, 2002, 74: 47-97

[32]　R Ferrer i Cancho. The variation of Zipf's law in human language, Eur Phys J B, 2005, 44: 249-257

[33]　Adamic L A, Huberman B. The nature of markets in the World Wide Web. Quarterly Journal of Electronic Commerce, 2000, 1: 5-12